# Towards Self-Optimizing Electron Microscope: Robust Tuning of Aberration Coefficients via Physics-Aware Multi-Objective Bayesian Optimization

Utkarsh Pratiush[1], Austin Houston[1], Richard Liu[1], Gerd Duscher[1], Sergei Kalinin[1]
[1]Department of Materials Science and Engineering, University of Tennessee, Knoxville, TN 37996, USA
* Corresponding author: upratius@vols.utk.edu

## Abstract

Realizing high-throughput aberration-corrected Scanning Transmission Electron Microscopy (STEM) exploration of atomic structures requires rapid tuning of multipole probe correctors while compensating for the inevitable drift of the optical column. While automated alignment routines exist, conventional approaches rely on serial, gradient-free searches (e.g., Nelder-Mead) that are sample-inefficient and struggle to correct multiple interacting parameters simultaneously. Conversely, emerging deep learning methods offer speed but often lack the flexibility to adapt to varying sample conditions without extensive retraining. Here, we introduce a Multi-Objective Bayesian Optimization (MOBO) framework for rapid, data-efficient aberration correction. Importantly, this framework does not prescribe a single notion of image quality; instead, it enables user-defined, physically motivated reward formulations (e.g., symmetry-induced objectives) and uses Pareto fronts to expose the resulting trade-offs between competing experimental priorities. By using Gaussian Process regression to model the aberration landscape probabilistically, our workflow actively selects the most informative lens settings to evaluate next, rather than performing an exhaustive blind search. We demonstrate that this active learning loop is more robust than traditional optimization algorithms and effectively tunes focus, astigmatism, and higher-order aberrations. By balancing competing objectives, this approach enables "self-optimizing" microscopy by dynamically sustaining optimal performance during experiments.

## I. Introduction

Scanning Transmission Electron Microscopy (STEM)(Pennycook, 2011; Williams et al., 2009; Crewe, 1974) has become one of the foundational tools across materials science, condensed matter physics, chemistry, catalysis, and related fields, enabling imaging and characterization of materials from the nanoscale down to atomic resolution and providing direct insight into atomic and molecular structure. As a result, STEM is now a primary instrument in academic and industrial research laboratories worldwide(Muller, 2009; Krivanek et al., 2010). Its versatility is further enhanced through integration with complementary techniques such as electron energy loss spectroscopy (EELS)(Egerton, 2011; Mkhoyan et al., 2007), which enables detailed analysis of chemical composition(Schneider et al., 1997; Williams & Carter, 1996), electronic structure, and low-energy quasiparticles. Together, atomic-scale imaging and spectroscopic capabilities make STEM a critical platform for elucidating structure–property relationships across diverse material systems and for driving the development of advanced technologies, including semiconductors(Nakamae, 2021), solar cells(Noircler et al., 2021), catalysts(Qu et al., 2023), and battery materials(Yu et al., 2021).

Despite these capabilities, the practical throughput of STEM remains severely bottlenecked by the manual and labour-intensive nature of instrument operation(Kalinin, Ophus, et al., 2022; Kalinin, Ziatdinov, et al., 2022; Liu et al., 2025; Pratiush et al., 2024). For image optimization, modern aberration-corrected scanning transmission electron microscopy (STEM) enables sub-angstrom imaging, yet maintaining optimal probe conditions remains a fragile and labour-intensive challenge(Erni, 2015; Kalinin et al., 2021). While hardware advances have pushed resolution limits, tuning multiple aberration coefficients is traditionally performed sequentially(Ishikawa et al., 2021) and assumes a single static imaging objective (Pattison et al., 2025; Ma et al., 2025a, 2025b). This process can take minutes, during which thermal drift, dose accumulation, and parameter coupling degrade performance, leaving no practical way to update the optical state dynamically during an experiment. While the classical human-based instrument optimization workflows made STEM the key tool for materials discovery, the fundamentally new opportunities in rapid materials discovery or atomic manipulation necessitate significant acceleration of the tuning tasks. To address this, researchers have explored a spectrum of automation strategies, ranging from classical control loops to deep learning.

Basic forms of automated alignment date back over 40 years. Early systems by Koike et al. (1974) and later Erasmus and Smith (1982)(Erasmus & Smith, 1982) utilized feedback loops to correct focus and astigmatism by optimizing "focus merit functions" derived from image shifts or contrast. Throughout the 1990s and 2000s, these evolved into standard routines that executed exhaustive focus sweeps to identify peak image quality metrics. With the advent of multipole correctors, methods shifted toward "blind" optimization directly on the sample of interest to avoid the interruptions required by tableau or Ronchigram methods. Kirkland(Kirkland, 2018, 2016) demonstrated via simulation that maximizing image variance could achieve optimal alignment. However, these "blind" search methods remain fundamentally limited by their serial nature and sample inefficiency. They typically optimize a single metric without regard for competing objectives, such as minimizing beam dose versus maximizing sharpness, and do not retain a memory of the parameter landscape.

More recently, machine learning has offered faster alternatives. Systems like DeepFocus(Schubert et al., 2024) and neural-network-based Ronchigram analysis(Bertoni et al., 2023) can infer aberrations in a single step. Yet, these Deep Learning approaches often require massive pre-training on specific datasets and struggle to adapt to novel imaging conditions or varying sample types. [Muller pt1 pt2 papers?]

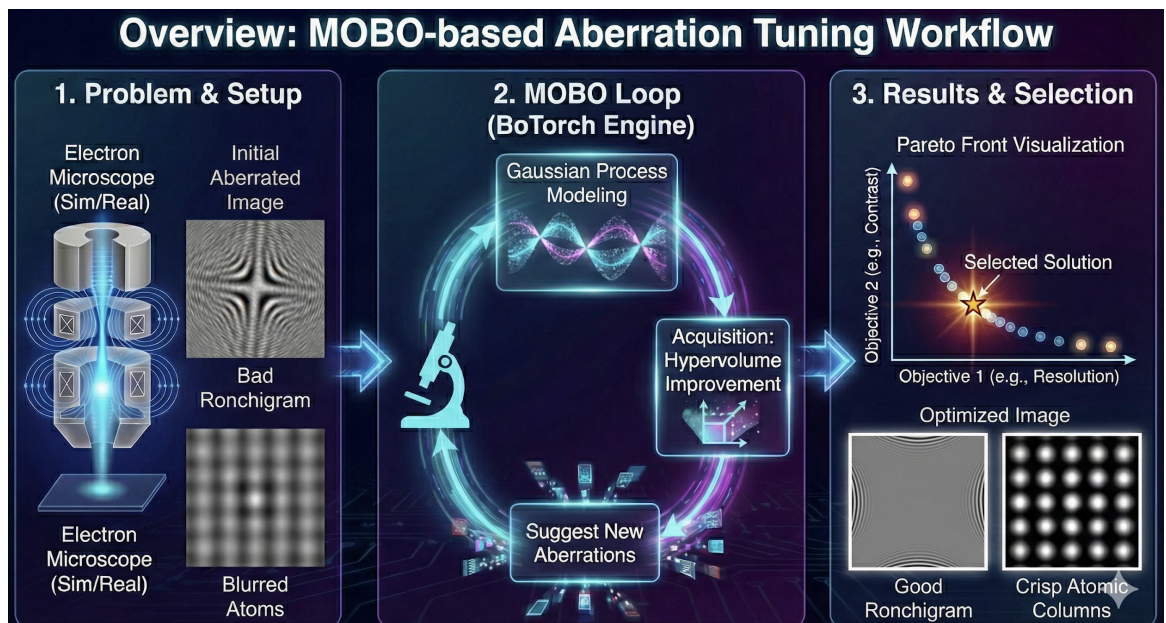

**Figure 1.** Overview of the multi-objective aberration-tuning workflow. Panel 1 (left): Initial probe formation and aberration parameterization. Panel 2 (center): Optimization loop combining image-quality metrics and multi-objective acquisition. Panel 3 (right): Final tuned

aberration state and corresponding image improvements. Figure created with assistance from an LLM.

We emphasize that the goal of this work is not to define a universal or optimal image-quality metric for aberration correction. Instead, different experimental contexts naturally motivate different, physically meaningful reward formulations, ranging from contrast and resolution to symmetry-induced objectives. By framing aberration correction as a multi-objective problem, Pareto fronts provide a principled way to expose and interpret trade-offs between these competing priorities, rather than collapsing them into a single scalar objective. Accordingly, the focus of this paper is on the optimization framework and the structure revealed by these trade-offs, rather than on advocating a specific reward function.

However, automated optimization of STEM imaging requires identification of the well-defined measure of image quality that can be used as target function for optimization. Even for atomically resolved imaging this choice can be ambiguous, and situation becomes even more complex for mesoscale imaging. To bridge the gap between slow iterative searches and rigid pre-trained models, we introduce a Multi-Objective Bayesian Optimization (MOBO) workflow (Figure. 1). Unlike memoryless search algorithms, our approach utilizes Gaussian Process modelling to intelligently explore the aberration space. By treating alignment as a multi-objective problem, we allow balancing multiple probabilistic definitions of image quality as competing hypotheses. The MOBO engine efficiently navigates the parameter space to identify the optimal Pareto front, which shape provides the measure of the robustness of reward functions and allows human in the loop interventions. This allows for "self-optimizing" microscopy that tunes aberrations rapidly and robustly, minimizing sample exposure while maximizing image quality.

## II. Methods

The fundamental challenge of aberration correction is the precise tuning of the electron probe to minimize phase errors. We define our search space $\mathcal{X}$, as the set of tunable aberration coefficients corresponding to the microscope's multipole correctors. This includes lower-order terms such as defocus $C_1$ and two-fold astigmatism $A_1$ as well as potentially higher-order terms ($B_2$, $A_2$, etc.) we refer the reader to(Ma et al., 2025a, 2025b) for detailed discussion. The goal of the optimization algorithm is to identify the parameter vector $x \in \mathcal{X}$, that yields the optimal imaging state.

## II. A. Aberrations Simulator and Synthetic Dataset

As a first step, we realize the MOBO optimization workflows using a physics-based simulation to generate a ground-truth dataset. This allows both for benchmarking the MOBO method and allows for known ground truth. The simulation environment is built upon the abtem (Madsen & Susi, 2021) and pyTEMlib (https://github.com/pycroscopy/pyTEMlib) frameworks, simulating images of a Tungsten Diselenide(WS$_2$) monolayer.

The optical model simulates a Thermo Fisher Spectra 300 instrument operating at 60 kV with a 30 mrad convergence angle, matching typical experimental parameters. We generate the electron probe wavefunction by explicitly defining a table of aberration coefficients including defocus ($C_{1,0}$), two-fold astigmatism ($C_{1,2a}$, $C_{1,2b}$), and higher-order terms ($C_{2,1}$, $C_{2,3}$), which serve as the tunable parameters for the optimization loop. The resulting image is formed via the convolution of the sample's scattering potential with the probe's Point Spread Function (PSF).

To ensure the optimization is robust to experimental realities (the "sim-to-real" gap), we do not train on pristine data; instead, we corrupt the calculated intensity, *I(r)* with two distinct noise layers. First, we inject spatially correlated Gaussian noise, filtered to emphasize low frequencies to mimic environmental instability, detector drift, and background fluctuations common in high-magnification scanning. Simultaneously, we apply Poissonian shot noise to the normalized image intensity, simulating the stochastic nature of electron detection at finite doses (e.g., 10^7 counts). This pipeline outputs a stream of "noisy" observations where the ground-truth optimal state (zero aberrations) is known, allowing us to rigorously quantify the convergence speed and precision of the Bayesian optimization algorithm before deployment on the physical instrument. The simulated environment is provided as an open code at
https://github.com/pycroscopy/asyncroscopy/blob/main/notebooks/Aberrations.ipynb.

## II.B. Defining Image Quality as Reward Functions

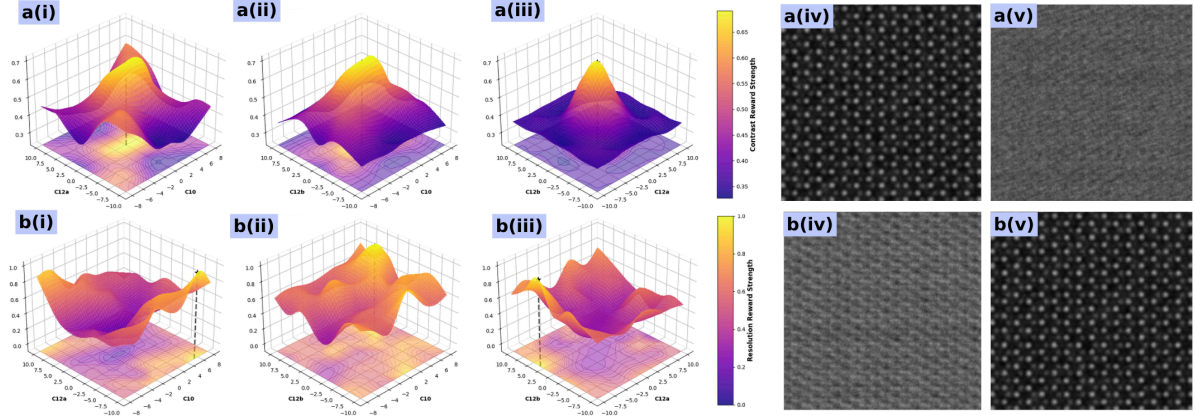

**Figure 2.** Multi-objective aberration landscape analysis and corresponding STEM image quality. (a) Contrast-based reward landscape as a function of aberration parameters C10, C12a, and C12b. Sub-panels a(i)–a(iii) show 3D surface plots for each 2D slice of the contrast reward field, highlighting the presence of local optima and the strong dependence of contrast on third-order aberrations. a(iv) and a(v) show representative simulated STEM images corresponding to high-contrast and low-contrast regions of the landscape. (b) Resolution-based reward landscape for the same aberration subspace. Sub-panels b(i)–b(iii) display the resolution-score surfaces, demonstrating a markedly different topology compared to the contrast objective, with distinct optimal regions in the aberration parameter space. b(iv) and b(v) show STEM images representing the best- and worst-resolution aberration settings. See also supplementary for similar plot for 2$^{nd}$ order aberrations.

In manual operation, human operators tune the corrector parameters by visually maximizing specific features of image quality, such as the "sharpness" of atomic columns or the contrast of the lattice. Often, these parameters are heuristic and in many cases optimizing for one parameter leads to the degradation of the others, resulting in strong operator bias and general lack of traceability in the principles by which instrument performance is optimized. As a necessary step towards automation of this process, these qualitative assessments must be translated into quantitative, scalar reward functions, $y = f(x)$ where yi is … and x is … .

Here, To ground our optimization in physics based metrics, we define a reward function based on two complementary metrics that serve as rigorous proxies for beam quality. First, we calculate the Normalized Image Variance (RMS Contrast), defined as the standard deviation of pixel intensity normalized by the mean image intensity. This normalization decouples contrast from total beam current, ensuring that high scores reflect a tightly focused probe rather than simple brightness fluctuations.

$$R_{\text{contrast}} = \frac{\sigma}{\mu + \epsilon} \qquad (1)$$

Where, σ and μ represent the standard deviation and mean of the image pixel intensities, respectively. ϵ is a small stability constant to prevent division by zero.

Alternatively, the reward function can by constructed by evaluating FFT Spectral Power to quantify crystallinity. Here, we apply a high-pass mask to the center of the frequency domain before calculating the mean spectral power; this eliminates uninformative background signals (the DC component), isolating the Bragg peaks that serve as a direct signature of atomic lattice resolution. It is worth noting that this reward regime represents a rich area for further exploration: by explicitly imposing crystallographic symmetries in both real and reciprocal space, one could physically constrain the search landscape to achieve significantly faster optimization.

$$R_{fft} = \langle \log(1 + |F|) \cdot M \rangle \qquad (2)$$

In the spectral objective $R_{fft}$, $\mathcal{F}$ denotes the 2D Fast Fourier Transform of the image, $|\cdot|$ extracts the magnitude, and $M$ is a binary high-pass mask that explicitly zeros out the central DC component to isolate lattice periodicity from background brightness, with $\langle \cdot \rangle$ indicating the arithmetic mean over the frequency domain. By treating the microscope output as a black-box function, the machine learning agent seeks to find $x^* = \arg\max f(x)$.

## II.C. Multi-Objective Optimization and the Pareto Front

A single metric is often insufficient to capture the full definition of "optimal" imaging; for instance, a setting that maximizes variance might overly emphasize noise, while one that maximizes FFT intensity might favor periodic artifacts. Therefore, we frame the tuning process as a Multi-Objective optimization problem. For more details on MOBO as a technique, please check the appendix.

Rather than scalarizing these rewards into a single weighted sum, we treat them as a vector of objectives, $f(x) = [f_1(x), f_2(x)]$ The goal of the MOBO algorithm is to approximate the Pareto Front, the set of solutions where no single objective can be improved without degrading another.

Deploying this multi-objective framework on live hardware offers critical advantages over single-metric approaches. First, it ensures experimental robustness by acting as a "contaminant filter"; by forcing the ML agent to maximize both contrast and crystallinity simultaneously, we prevent "reward hacking" where the optimizer might otherwise lock onto high-contrast artifacts (like dirt or surface damage). Second, it facilitates objective correlation analysis via the Pareto front, allowing us to empirically quantify fundamental trade-offs in image formation physics. For example, if different heuristic reward functions are maximized in a single region of the control parameter space, that means that there is a single optimal imaging condition. At the same time if Pareto front has finite extent, this implies that rewards are misaligned but then this structure supports a "human-in-the-loop" workflow, empowering domain experts to intervene and select the optimal trade-off from the generated candidates.

## II. D. Alignment with FAIR Data Principles

Finally, our framework adheres to FAIR (Findable, Accessible, Interoperable, Reusable) principles by treating the optimization trajectory itself as a high-value dataset, rather than a transient operational step. Unlike manual tuning where operators discard suboptimal images, the MOBO engine logs every iteration, linking the applied aberration coefficients (Action) to the resulting image and reward metrics (Observation). This ensures the data is Findable and Accessible for post-analysis, while the use of fundamental physics-based rewards (e.g., FFT based reward) ensures the results are Interoperable across different microscope platforms. Crucially, this persistent record allows even "failed" alignment steps to be Reusable for training future surrogate models or digital twins, ensuring that every second of beam time contributes to a lasting, cumulative knowledge base. When implemented on a microscope, the alignment trace also preserves the information on the timing of each operation, providing the way to estimate the instrument time and matching required computational resources.

## II E. On Real microscope

All experiments were conducted on a ThermoFisher Scientific Spectra 300 transmission electron microscope. Aberration lenses were tuned using the CEOS API. Additional instrument control (stage and detectors) was managed via the ThermoFisher AutoScript API. To validate the optimization framework, we utilized a standard resolution test sample consisting of gold (Au) nanoparticles.

The computational workload was offloaded to high-performance computing (HPC) resources accelerated by NVIDIA A6000 GPUs. The core optimization logic was implemented using BoTorch for the Bayesian Optimization loop, with Gaussian Process surrogates modeled via GPyTorch. Comprehensive experimental logging was maintained using Python's standard logging libraries to ensure full reproducibility of the decision pathways.

The code to orchestrate the hardware used in this work evolved in two distinct phases. Initially, the CEOS control interface was developed and deployed as an integrated module within the stemOrchestrator(Pratiush et al., 2025, 2026) (https://github.com/pycroscopy/pyAutoMic/tree/main/TEM/stemOrchestrator) framework, which generated the early validation results presented here. In subsequent developments aimed at enabling distributed asynchronous workflows, this interface was refactored and inculcated as a standalone server within the asyncroscopy(https://github.com/pycroscopy/asyncroscopy/tree/main/asyncroscopy) package. Each component which led to the successful deployment of the workflow on the instrument is available opensource at provided links.

## III Results and discussion

### III a. Pareto front on the grid data

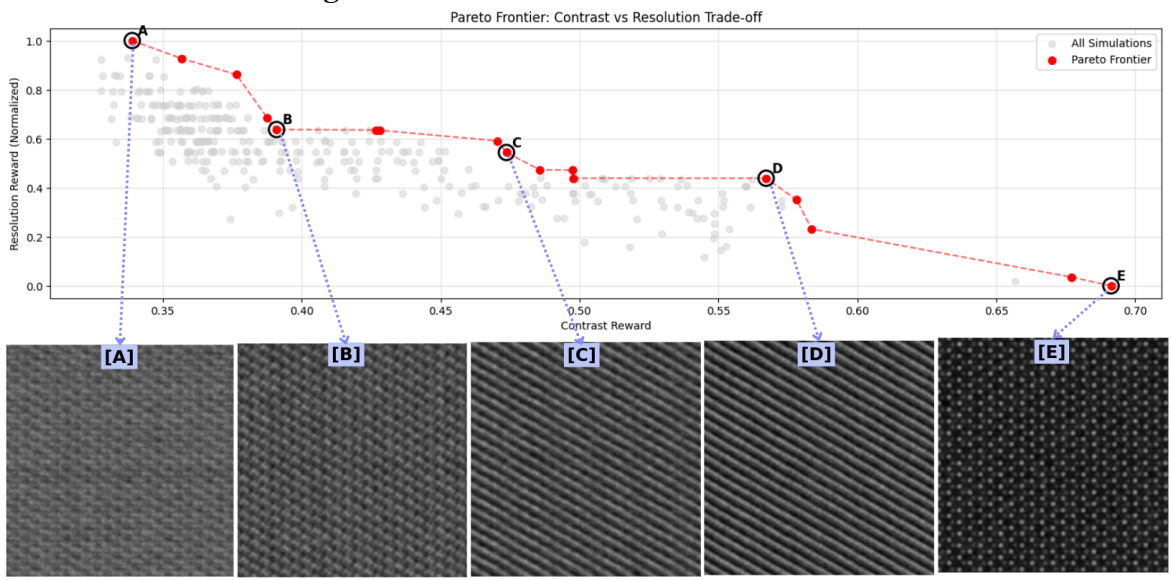

**Figure 3.** The scatter plot shows all simulated aberration configurations evaluated in the 1st-order aberrations (defocus $C_1$ and two-fold astigmatism $A_1$) space, with each point representing a pair of (contrast reward, resolution reward). The Pareto frontier (red) identifies the set of non-dominated solutions where improvement in contrast necessarily incurs a loss in resolution, and vice-versa. Five representative Pareto points (A–E) are highlighted to illustrate distinct regions of the trade-off surface. See also supplementary for similar plot for 2nd order aberrations.

To obtain a fundamental understanding of the reward landscape and the geometry of the trade-offs, we conducted an exhaustive grid search over the 1st-order aberration space within the simulation environment. This dense sampling allows us to empirically map the distribution of all possible solutions and explicitly identify the global Pareto frontier, verifying the conflict between metrics independent of any specific optimization trajectory.

The scatter plot in Figure 3 reveals that the search space is overwhelmingly dominated by a dense "grey cloud" of suboptimal solutions. These interior points represent states of deep misalignment where multiple aberration coefficients are simultaneously incorrect, yielding images that lack both contrast and resolution. The sparse population of points along the red Pareto frontier illustrates that the optimal solution is not a broad basin, but a thin, sensitive manifold at the edge of a high-dimensional error landscape.

This distribution empirically demonstrates the wastefulness of traditional grid searches, which allocate beam time uniformly across the "grey cloud" rather than refining the optimal boundary. While feasible for three parameters, this approach scales disastrously with higher-order aberrations due to the curse of dimensionality ($N^d$). As the parameter space expands, the probability of a fixed grid landing on the narrow Pareto frontier approaches zero, rendering exhaustive search mathematically intractable for realistic microscope tuning.

The labelled points (A–E) in Figure 3, expose the danger of optimizing a single metric. Point A maximizes the "Resolution" reward but produces a noisy, artifact-ridden image, while Point E maximizes "Contrast" but can sacrifices fine spatial detail(see figure 5). A single-objective ML agent would blindly converge to these extremes ("reward hacking"), effectively failing the experiment. Only the multi-objective approach identifies the balanced "sweet spot" (Points C and D), confirming that robust optical alignment requires navigating the trade-off between competing physical objectives.

The necessity for an intelligent optimization strategy becomes evident when analyzing the complexity of the parameter space. Even with a relatively coarse discretization, we identify approximately 343 candidate states for 1st-order aberrations and 256 for 2nd-order terms only; however, denser grid causes the volume of candidates to increase according to a power law, rendering exhaustive search methods intractable. Consequently, we employ Bayesian Optimization to efficiently navigate this expanding manifold and identify optimal candidates with minimal sampling. To validate this approach, we utilize a high-fidelity simulator which serves a critical dual purpose: it allows us to map the reward landscape and Pareto front across a dense grid, an exercise that would be prohibitively expensive and time-consuming on a physical instrument while simultaneously providing a controlled environment to rigorously decouple and understand the underlying physics of the aberrations.

**III b. Pareto front using BO in simulated environment**

Prior to experimental deployment, we validated the agent's behavior in a controlled simulation environment. This step was critical to verify that the physics-based reward functions correctly mapped the aberration landscape without the confounding factors of environmental noise.

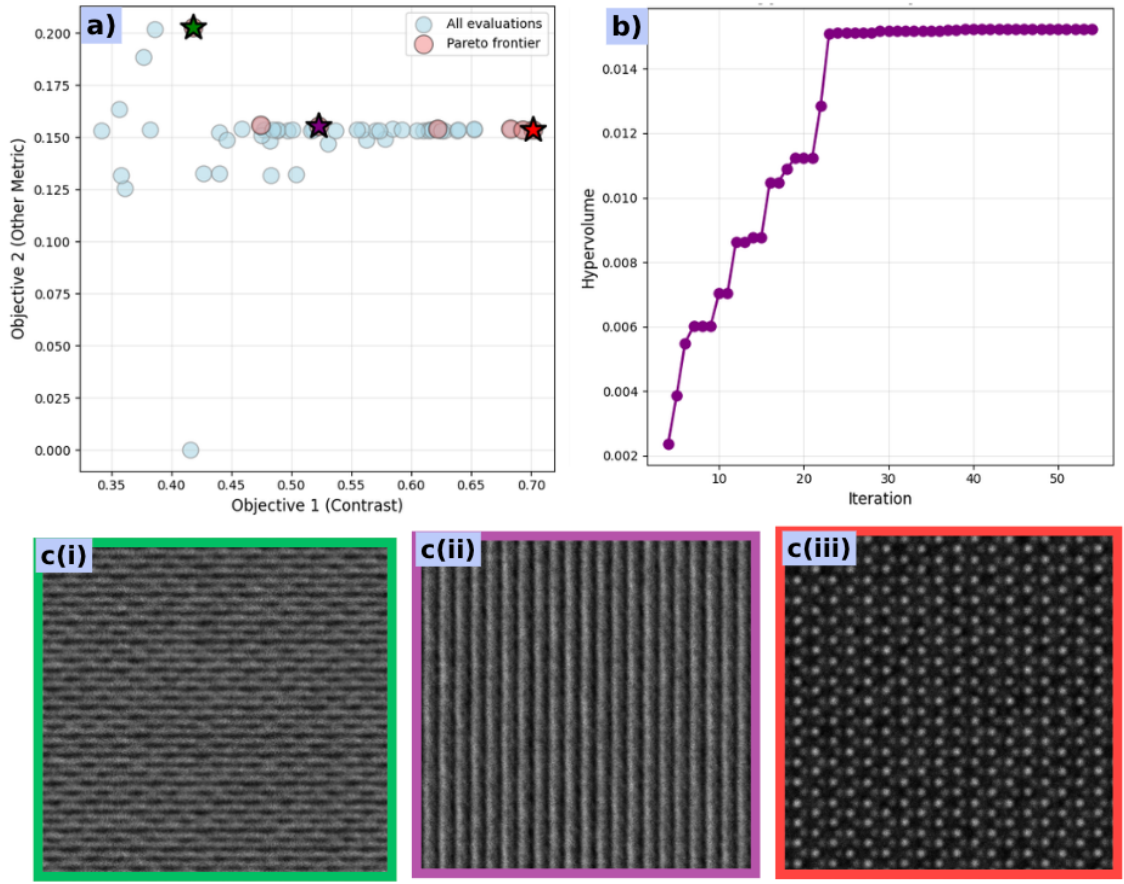

**Figure 4.** Validation of the multi-objective Bayesian optimization workflow in a simulated STEM environment for 1st-order aberration correction (Defocus $C_1$ and 2-fold Astigmatism $A_1$). (a) The objective space illustrates the trade-off landscape between Contrast reward and FFT reward. The Pareto frontier (red circles) marks the most efficient solutions discovered. (b) Hypervolume metric versus iteration, indicating convergence to the optimal solution set within approximately 25 iterations. (c) Representative simulated micrographs corresponding to the starred states in (a): (i) A state with high spectral periodicity but low contrast (Green star); (ii) A strongly astigmatic state showing characteristic directional elongation of the atomic columns (Purple star); and (iii) The Pareto-optimal solution (Red star) which simultaneously maximizes contrast and spectral power to yield pristine, atomically resolved lattices.

Figure 4 summarizes the optimization trajectory. As observed in the objective space in Figure 4(a), the agent successfully identifies the Pareto frontier, effectively navigating the trade-off between image contrast and lattice periodicity. The representative micrographs in Figure 4(c) explicitly visualize the physical distortions associated with the search path. The solution marked by the Purple star (c(ii)) captures a local optimum dominated by astigmatism, evidenced by the directional elongation ("streaking") of the atomic columns. However, the optimizer successfully escapes this astigmatic manifold, converging to the global optimum marked by the Red star (c(iii)). Here, the agent achieves simultaneous maximization of both contrast and FFT spectral power, producing a sharp, aberration-free image that serves as the ground-truth baseline for our subsequent hardware experiments.

**III c. MOBO on Real Microscope**

Before moving from simulated dataset to real instrument we must also account for systematic discrepancies between the idealized simulation and the physical instrument. In a real microscope, hardware components introduce non-idealities that directly affect reward calculation. For instance, the operation of the beam blanker during scan flyback can introduce high-frequency periodic artifacts or dampen specific Fourier coefficients, causing the experimental FFT intensity to differ from the theoretical scattering potential. Similarly, hysteresis in the magnetic lenses or non-linearities in the scan coils can distort the perceived lattice symmetry. While we do not explicitly model every hardware artifact, the injection of heavy correlated noise serves as a surrogate for these unmodeled perturbations, forcing the ML agent to learn robust policies rather than overfitting to perfect simulator physics.

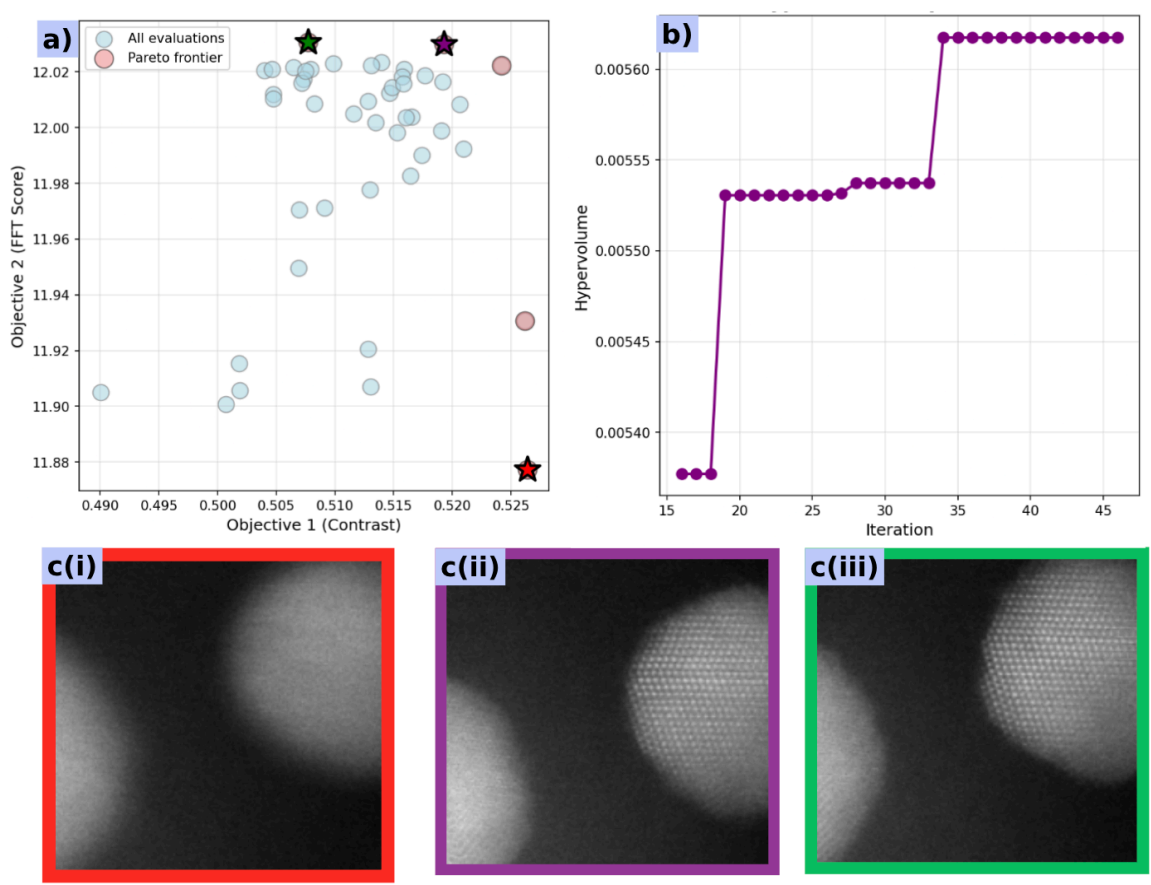

**Figure 5.** Experimental deployment of the multi-objective optimization on the ThermoFisher Spectra 300 for simultaneous tuning of defocus, 1st and 2nd order aberrations(7 Dimensional space). (a) The experimental objective space reveals a distinct Pareto front (red circles) characterizing the trade-off between Contrast reward and FFT reward. (b) Hypervolume improvement confirming robust convergence on the physical instrument. (c) Representative experimental micrographs corresponding to the starred solutions in (a): (i) A high-contrast but defocused state (Red star), illustrating the risk of single-objective optimization; (ii) a state with improved lattice visibility (Purple star); and (iii) the Pareto-optimal solution (Green star)

where the method balances contrast and resolution to resolve sharp atomic lattices on the Au nanoparticles.

Following the simulation benchmarks, we deployed the optimization directly on the physical microscope column to tune 1st and 2nd order aberrations (defocus and astigmatism) on a standard Au nanoparticle sample. Figure 5 summarizes this real-world performance, offering empirical validation of our "physics-aware" reward design.

The objective space in Figure 5(a) reveals a critical physical insight: the state of maximum contrast is *not* necessarily the state of maximum resolution. The solution marked by the Red star, corresponding to image c(i), achieves the highest contrast score (Objective 1 ~ 0.525) yet fails to resolve the atomic lattice (Objective 2 < 11.90). This confirms our hypothesis that single-metric optimization can be misleading; an AI driven solely by contrast might lock onto this defocused, high-signal state.

By utilizing the multi-objective Pareto front, the model successfully navigated away from this local optimum. As the Hypervolume increases in Figure 5(b), the optimizer discovers the trade-off curve, eventually identifying the solution marked by the Green star. As shown in d(iii), this state sacrifices a marginal amount of global contrast to maximize spectral power, resulting in the successful resolution of distinct atomic columns. This ability to autonomously distinguish between "bright/blurry" and "sharp/crystalline" demonstrates the robustness of the framework against the noise and ambiguity inherent in live experiments.

**III.d. Optimization cost function and instrumentation co-design**

To effectively implement automated experimentation, it is critical to analyze the time-cost breakdown of the optimization loop. By monitoring the wall-clock time for every step of the process, we can disentangle the fixed costs of hardware interaction from the scaling costs of the computational algorithm.

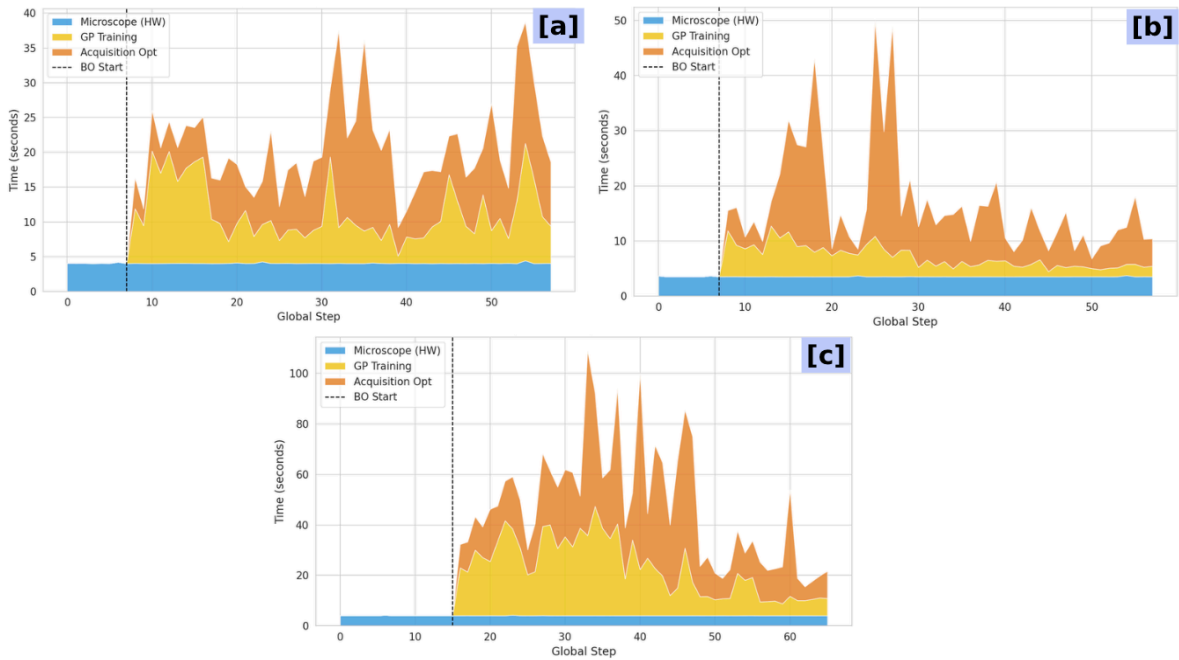

**Figure 7.** Computational vs. Hardware cost breakdown across increasing search space dimensionalities. Stacked area charts displaying the time elapsed (in seconds) per global step for three distinct optimization scenarios: (a) represents 3-parameter search space

(configuration c1-a1); (b) shows 4-parameter search space (configuration b2-a2); and (c) has 7-parameter combined search space (c1-a1-b2-a2). The vertical dashed line indicates the start of the Bayesian Optimization (BO) phase. The total time is stratified into hardware communication latency (Microscope (HW), blue) and computational overhead, consisting of GP Training (yellow) and Acquisition Optimization (orange). While hardware latency remains constant ($\approx$ 4s) across all regimes, computational cost scales super-linearly with dimensionality, becoming the dominant bottleneck in the 7-parameter case.

A detailed timing breakdown for Bayesian Optimization (BO) runs across three different search space dimensionalities on the real microscope is shown in Figure 7. The total time per step is decomposed into three components: Microscope Hardware (HW) interaction time (blue), Gaussian Process (GP) Training (yellow), and Acquisition Function Optimization (orange). A distinct trend emerges when comparing low-dimensional searches to high-dimensional ones. In lower dimensions (e.g., 3 parameters in Panel a), the computational overhead is comparable to the hardware latency. However, as the dimensionality increases to 4 parameters (Panel b) and eventually 7 parameters (Panel c), the computational load specifically the GP training and acquisition optimization dominates the total iteration time. While the hardware operation remains a fixed "floor" (approximately 4 seconds), the compute time scales significantly, peaking at over 100 seconds per step in the 7-parameter case.

This timing analysis serves as a blueprint for "optimization-instrumentation co-design," guiding strategic resource investment based on the dimensionality of the problem. In compute-limited regimes (high $D$), where complex searches create algorithmic bottlenecks, investments yield the highest returns when focused on accelerating compute, such as utilizing GPU acceleration for GP regression or adopting more efficient acquisition strategies. Conversely, in instrument-limited regimes (low $D$), the algorithm is sufficiently fast that the bottleneck shifts to the physical apparatus, meaning efforts should prioritize reducing communication latency or increasing hardware throughput. By quantifying these distinct cost components, we can dynamically adapt our experimental strategy, ensuring that neither the computational overhead nor the instrument's physical limitations unnecessarily throttle the discovery process.

**IV Code availability(all the links):**

Aberrations simulator :
https://github.com/pycroscopy/asyncroscopy/blob/main/notebooks/Aberrations.ipynb

stemOrchestrator :
https://github.com/pycroscopy/pyAutoMic/tree/main/TEM/stemOrchestrator

Workflow code to try on live instrument :
https://github.com/pycroscopy/asyncroscopy/tree/main/notebooks/aberrations-BO

**V Conclusion**

We have demonstrated a Multi-Objective Bayesian Optimization (MOBO) framework that automates the complex task of aberration correction in scanning transmission electron microscopy. By replacing inefficient "blind" grid searches and gradient-free heuristics with a sample-efficient Gaussian Process model. We show that this approach enables a mechanism to tune the hardware based on physics based rewards and also diagnose the state of the beam.

Crucially, this framework addresses the fundamental mismatch between idealized theoretical rewards and physical instrument reality. Real-world microscopes suffer from non-idealities, such as beam blanker artifacts, hysteresis, and detector non-linearities that often cause single-objective optimizers to converge on fragile or artifact-ridden states (e.g., the "resolution trap"). MOBO compensates for these unmodeled deviations not by forcing a single solution, but by revealing the Pareto frontier of achievable performance. This makes the inevitable compromise between competing objectives (e.g., contrast vs. resolution) explicit and open for post-examination, empowering the human operator to select a robust operating point that balances theoretical optima with experimental stability.

Finally, this capability paves the way for advanced beam engineering, where the electron probe is not merely "corrected" to a round point but intentionally shaped to enhance sensitivity to specific material properties. By systematically tuning aberrations to break probe symmetry, future workflows could selectively maximize resolution along specific crystallographic directions. This would enable the precise mapping of lattice strains and subtle distortions that are currently difficult to resolve with standard circularly symmetric probes. MOBO serves as the critical enabler for this next generation of microscopy, providing the high-dimensional navigation required to engineer these complex, task-specific wavefunctions.

## IV Acknowledgement

(U. P, A.H, G. D, S.V.K) acknowledges support from high performance computing facility, ISAAC and Microscopy Core Facility, of The University of Tennessee, Knoxville (UTK). This work was partially supported (AH, GD) by the U.S. DOE, Office of Science, Materials Sciences and Engineering Division and the Center for Nanophase Materials Sciences, which is a DOE Office of Science User Facility. The work was partially supported (SVK) by the U.S. Department of Energy, Office of Science, Office of Basic Energy Sciences as part of the Energy Frontier Research Centers program: CSSAS—the Center for the Science of Synthesis Across Scales. We gratefully acknowledge the technical support provided by Heiko Müller and Ingo Massmann from CEOS GmbH, whose assistance was instrumental in establishing the robust hardware connections for the aberration corrector.

**Supplementary**

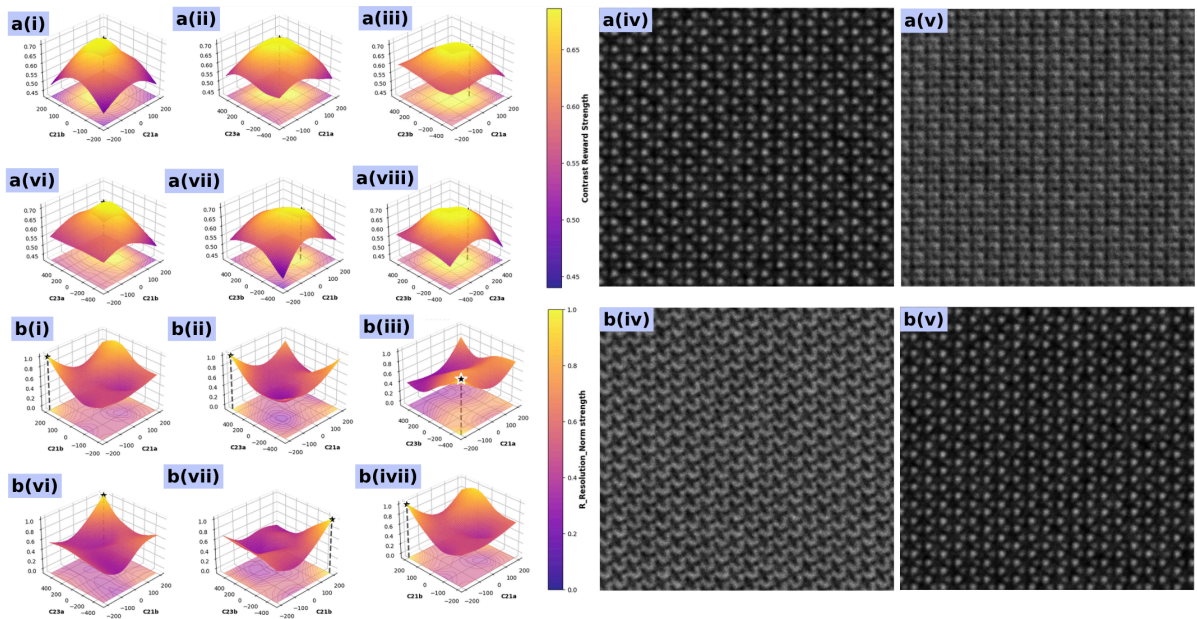

**Figure S1**. Contrast and resolution reward landscapes for higher-order aberrations (C21a, C21b, C23a, C23b) and associated STEM image quality. **(a)** Contrast reward surfaces. Sub-panels a(i)–a(iii) and a(vi)–a(viii) show contrast reward landscapes evaluated across pairs of higher-order aberrations, including (C21a, C21b) and (C23a, C23b) combinations. The landscapes exhibit smooth, dome-like topologies, indicating that contrast varies more gradually in this higher-order subspace compared to the strong nonlinear behavior observed for C10/C12 terms. The global maximum region is marked with a star. a(iv) and a(v) present representative simulated worst and best STEM images sampled from this reward surface. **(b)** Resolution reward surfaces. Sub-panels b(i)–b(iii) and b(vi)–b(viii) visualize the corresponding resolution reward fields across the same aberration combinations. Unlike contrast, resolution exhibits sharper gradients and distinct optimal "ridges," indicating higher sensitivity to C21 and C23 terms. The optimal region is denoted by a star. b(iv) and b(v) show example STEM images associated with the highest- and lowest-resolution reward values.

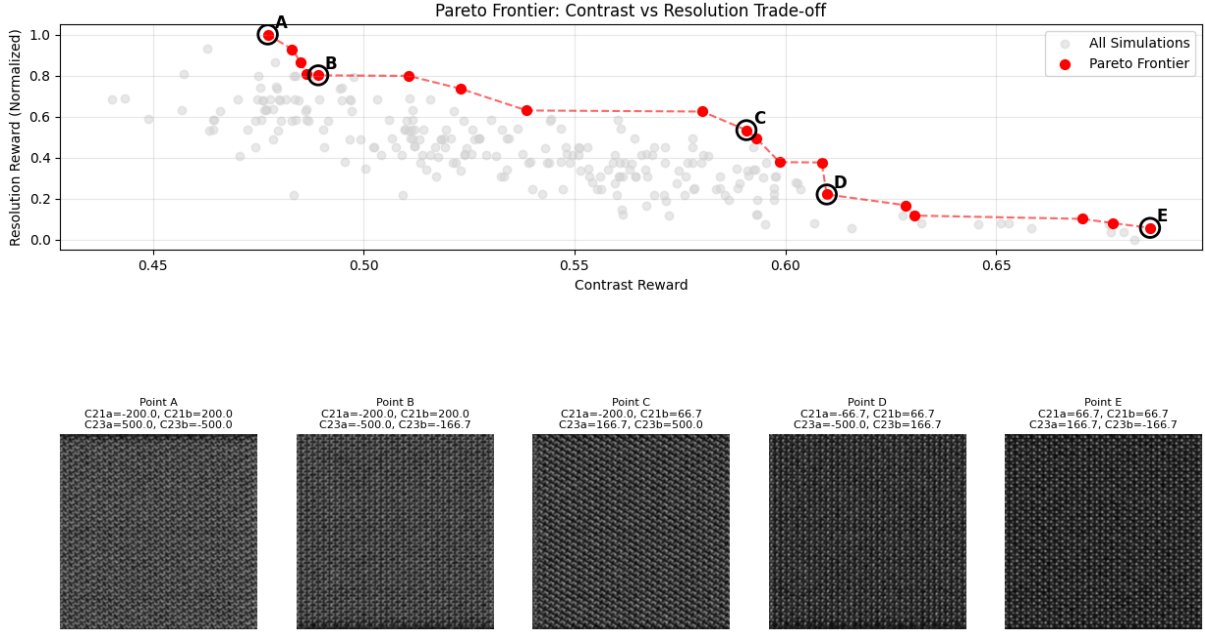

**Figure S2.** The scatter plot displays all simulated imaging conditions evaluated across the four-dimensional higher-order aberration space (**C21a**, **C21b**, **C23a**, **C23b**), with each point representing a pair of (contrast reward, resolution reward). The **Pareto frontier** (red) highlights the set of non-dominated solutions where improving contrast inevitably reduces resolution, and vice versa. Five frontier points (**A–E**) are marked to illustrate representative sections of the trade-off curve.

### II c. Bayesian Optimization with Surrogate modelling

### II c. i. Gaussian Process
A Gaussian Process (GP)(Rasmussen, 2004; Gardner et al., 2018) is a non-parametric probabilistic model that defines a distribution over possible functions. Instead of assuming a rigid functional form, a GP says: *any set of function values should jointly behave like a multivariate Gaussian*. This makes GPs flexible and well-suited for modelling smooth but unknown response surfaces.

A GP is fully defined by a mean function $m(x)$ and a covariance (kernel) function $k(x, x')$:

$$f(x) \sim GP\big(m(x), k(x, x')\big)$$

In practice, the mean is often taken as $m(x) = 0$, and smoothness is encoded through kernels such as the squared-exponential (RBF):

$$k(x, x') = \sigma^2 \exp\exp\left(-\frac{|x - x'|^2}{2l^2}\right)$$

where *l* controls how quickly the function can vary and $\sigma^2$ sets its overall scale.
**Intuition:** points that are close in input space should yield similar outputs—this is exactly what the kernel enforces.

Given training data $D = \{X, y\}$, the GP provides a **predictive posterior** at any new input $x_*$:

$$f(x_*) \mid D \sim N(\mu_*, \sigma_*^2)$$

with closed-form expressions:

$$\mu_* = k_*^\top (K + \sigma_n^2 I)^{-1} y, \quad \sigma_*^2 = k(x_*, x_*) - k_*^\top (K + \sigma_n^2 I)^{-1} k_*$$

Here,

$\mu_*$ is the predicted function value (the model's "best guess"),

$\sigma_*^2$ quantifies uncertainty, increasing where data are sparse,

and $K$ is the kernel matrix computed over the training inputs.

This combination of prediction and uncertainty makes GPs particularly valuable as surrogate models in Bayesian optimization.

### II c. ii. Multi Objective Gaussian Processes (MOGP)

In multi-objective(Daulton et al., 2020, 2021) settings, it is common to model each objective $f_i(x)$ using an independent Gaussian Process:

$$f_i(x) \sim GP(m_i(x), k_i(x, x')), i = 1, \ldots, m.$$

This yields a factorized joint posterior,

$$p(f(x) \mid D) = \prod_{i=1}^{m} p(f_i(x) \mid D_i),$$

while multi-objective acquisition functions (e.g., based on hypervolume improvement) operate on joint samples

$$f^{(s)}(x) = (f_1^{(s)}(x), \ldots, f_m^{(s)}(x)),$$

drawn simultaneously from these independent surrogates.

Although this independent-GP formulation is widely adopted due to its simplicity and scalability, multitask Gaussian Processes(Bonilla et al., 2007)—which incorporate cross-objective covariance through a task kernel—provide a richer joint model. We do not use multitask GPs in this work, but they may offer advantages when objectives are strongly correlated and represent an interesting direction for future exploration.

### II c. iii. Bayesian Optimization (BO)

Bayesian Optimization (BO)(Garnett, 2023; Balandat et al., 2019) is a sample-efficient strategy for optimizing functions that are expensive or noisy to evaluate. BO constructs a probabilistic surrogate model $p(f(x) \mid D)$ of the unknown objective and uses this model to guide the selection of the next evaluation point. At iteration $t$, BO chooses

$$x_{t+1} = \arg\max_x \alpha(x),$$

where $\alpha(x)$ is an acquisition function that balances **exploration** (sampling where uncertainty is high) and exploitation (sampling where predicted performance is favorable).

In the multi-objective case, the surrogate consists of independent GPs for each objective,

$$f_i(x) \sim GP(m_i(x), k_i(x, x')),$$

and the acquisition function—such as Expected Hypervolume Improvement (EHVI)—evaluates the joint predictive distribution to determine how much a new sample is expected to improve the Pareto front.

The key intuition behind BO is that the surrogate provides both a mean prediction and an uncertainty estimate, enabling the optimizer to focus evaluations on points that are most informative. This makes BO particularly effective in experimental workflows where each measurement is costly, time-consuming, or physically constrained.

### II c. iv. EHVI acquisition function

Active learning for multi-objective(Daulton et al., 2020; Balandat et al., 2019; Daulton et al., 2021) optimization requires selecting the most informative next experiment. In our setting, "informativeness" naturally arises from the concepts of Pareto optimality and acquisition functions that quantify expected improvement.

In multi-objective settings we aim to optimize a vector-valued function $f(x) = [f_1(x), f_2(x), \ldots, f_m(x)]$, with objectives being conflicting. A solution $x*$ is said to be *Pareto optimal* if no other solution exists that improves at least one objective without worsening another. Hence, a point x dominates another point x' if:

$$\forall i \in \{1, \ldots, m\}, f_i(x) \leq f_i(x') \quad \text{and} \quad \exists j, f_j(x) < f_j(x')$$

All non-dominated solutions together form the Pareto front, a frontier summarizing the trade-offs across objectives.

Expected Hypervolume Improvement (EHVI) is a Bayesian optimization acquisition function tailored to multi-objective settings. It measures the expected increase in dominated hypervolume obtained by evaluating a new candidate point $x$. The hypervolume itself represents the volume of objective space dominated by the current Pareto front relative to a chosen reference point.

For a current Pareto set $P$, the hypervolume improvement from evaluating a new point $f(x)$ is:
$$\Delta HV(x) = HV(P \cup \{f(x)\}) - HV(P)$$
Since $f(x)$ is not observed beforehand, we take the expectation with respect to the surrogate model's posterior (e.g., Gaussian Process), giving:
$$EHVI(x) = E_{f(x)}[\Delta HV(x)]$$
By selecting the candidate $x$ with the largest expected increase in dominated hypervolume, EHVI guides experimentation toward regions that most effectively enlarge the Pareto front. This criterion is especially effective in conjunction with probabilistic surrogates like Gaussian Processes, where uncertainty information can be fully exploited.